\documentclass[review, 12pt]{elsarticle}

\usepackage{amssymb}
\usepackage{amsmath}

\usepackage{color}
\usepackage{url}
\usepackage{graphicx}
\usepackage{subfigure}
\usepackage{multirow}
\usepackage{hyperref}
\usepackage{amsmath}

\usepackage{graphicx}
\usepackage{color}
\usepackage{multirow}
\usepackage{tabularx}
\usepackage{array}
\usepackage{mathrsfs}
\usepackage{amssymb}

\usepackage{amssymb,amsmath,enumerate,threeparttable,bm,subfigure,graphicx}
\usepackage{amsmath, amsfonts}
\usepackage{algorithm}
\usepackage{algorithmic}

\journal{Artificial Intelligence in Medicine}

\begin{document}

\begin{frontmatter}



\title{Robust Source-Free Domain Adaptation for Medical Image Segmentation based on Curriculum Learning} 


\author[a,b]{Ziqi Zhang}
\author[c]{Yuexiang Li}
\author[d]{Yawen Huang}
\author[d]{Nanjun He}
\author[b]{Tao Xu}
\author[e]{Liwei Lin} 
\author[f]{Yefeng Zheng}
\author[a]{Shaoxin Li\corref{cor1}}
\ead{lsx30716@rjh.com.cn}
\author[a]{Feiyue Huang\corref{cor1}}
\ead{hfy30711@rjh.com.cn}
\cortext[cor1]{Corresponding authors}


\affiliation[a]{organization={Shanghai Digital Medicine Innovation Center, Ruijin Hospital},
            city={Shanghai},
            postcode={200025}, 
            country={China}}

\affiliation[b]{organization={Tsinghua University},
            city={Shenzhen},
            postcode={518057}, 
            country={China}}

\affiliation[c]{organization={Medical AI ReSearch (MARS) Group, Guangxi Key Laboratory for Genomic and Personalized Medicine, Guangxi Medical University},
            city={Nanning},
            postcode={530021}, 
            country={China}}
            
\affiliation[d]{organization={Jarvis Research Center, Tencent Youtu Lab},
            city={Shenzhen},
            postcode={518000}, 
            country={China}}

\affiliation[e]{organization={University of California, Berkeley},
            city={Berkeley},
            country={United States}}

\affiliation[f]{organization={Westlake University},
            city={Hangzhou},
            postcode={310024}, 
            country={China}}

\begin{abstract}
Recent studies have uncovered a new research line, namely source-free domain adaptation, which adapts a model to target domains without using the source data. Such a setting can address the concerns on data privacy and security issues of medical images. However, current source-free domain adaptation frameworks mainly focus on the pseudo label refinement for target data without the consideration of learning procedure. Indeed, a progressive learning process from source to target domain will benefit the knowledge transfer during model adaptation. To this end, we propose a curriculum-based framework, namely learning from curriculum (LFC), for source-free domain adaptation, which consists of easy-to-hard and source-to-target curricula. Concretely, the former curriculum enables the framework to start learning with `easy' samples and gradually tune the optimization direction of model adaption by increasing the sample difficulty. While, the latter can stablize the adaptation process, which ensures smooth transfer of the model from the source domain to the target. We evaluate the proposed source-free domain adaptation approach on the public cross-domain datasets for fundus segmentation and polyp segmentation. The extensive experimental results show that our framework surpasses the existing approaches and achieves a new state-of-the-art.

\end{abstract}



\begin{highlights}
\item We propose a framework, learning from curriculum (LFC), for source-free domain adaptation.
\item Our LFC consists of novel two curricula---easy-to-hard and source-to-target.
\item The proposed LFC achieves a new state-of-the-art on publicly available datasets.

\end{highlights}



\end{frontmatter}




\section{Introduction}
Unsupervised domain adaptation (UDA) has proven to be very effective on reducing the domain discrepancy.  
However, conventional UDA approaches require the source data to learn a domain invariant feature representation for the target domain. Such a setting may not be guaranteed in many real-world scenarios, due to the data privacy and security issues.
To tackle the challenge, recent studies \cite{liu2021source,yansource,chen2021source} uncovered an interesting-yet-challenging research line, namely source-free unsupervised domain adaptation. 
Specifically, researchers are encouraged to develop adaptation methods only using the source-domain-trained model and unlabeled target data. Due to the lack of source data, existing UDA methods \cite{8764342,Chen_Dou_Chen_Qin_Heng_2019,Xie_2020_MICCAI,Xue_2020_MICCAI,Chen_TM_2021} failed to deal with such a challenging setting. In this regard, several specific source-free domain adaptation methods \cite{li2020model,kurmi2021domain,chen2021source,YANG2022102457,SFD-MM,MIIP-2022} have been proposed. For example, Li \emph{et al.} \cite{li2020model} utilized conditional GANs to generate labeled target data and finetune the source model with multiple semi-supervised model regularization terms. Kurmi \emph{et al.} \cite{kurmi2021domain} used conditional generative adversarial networks (GANs) to simultaneously perform data generation and domain adaptation. Kim \emph{et al.} \cite{kim2020towards} exploited pseudo labels in the target domain to improve the model generalization performance. Dou \emph{et al.} \cite{chen2021source} presented a denoised pseudo-labeling method, introducing two complementary pixel-level and class-level denoising schemes with uncertainty estimation and prototype estimation to reduce noisy pseudo labels. However, the existing frameworks primarily focused on the pseudo label refinement, which lacked consideration of the learning procedure. Indeed, researchers \cite{9461766,9392296,Wei_2021_WACV} have validated that a progressive learning process from easy to hard samples (known as {\itshape curriculum learning}) will benefit the feature representation learning of deep learning models.

In this paper, we propose a curriculum-based framework, namely learning from curriculum (LFC), for source-free domain adaptation, which consists of two curricula---easy-to-hard and source-to-target. In particular, the former curriculum enables the model to initially pay more attentions on the `easy' target data, which locates in the mutual space of the source and target domains in the latent space, and gradually vary its focus to the `hard' samples during source-free adaptation. While, the latter one, formed by our smooth update strategy, can stabilize the adaptation process and ensures smooth transfer of the model from source domain to the target.
To further boost the domain adaptation performance, we integrate a self-supervised task into source-to-target curriculum. Concretely, the model is enforced to generate consistent predictions with different data augmentations (\emph{e.g.,} photometric noise, flipping and scaling), which enables it to exploit transformation-invariant features from unlabeled target data. The proposed LFC framework is evaluated on publicly available cross-domain datasets for fundus segmentation and polyp segmentation, respectively.
The extensive experimental results show that our framework surpasses existing approaches and achieves a new state-of-the-art.

\begin{figure}[!tb]
    \includegraphics[width=\textwidth]{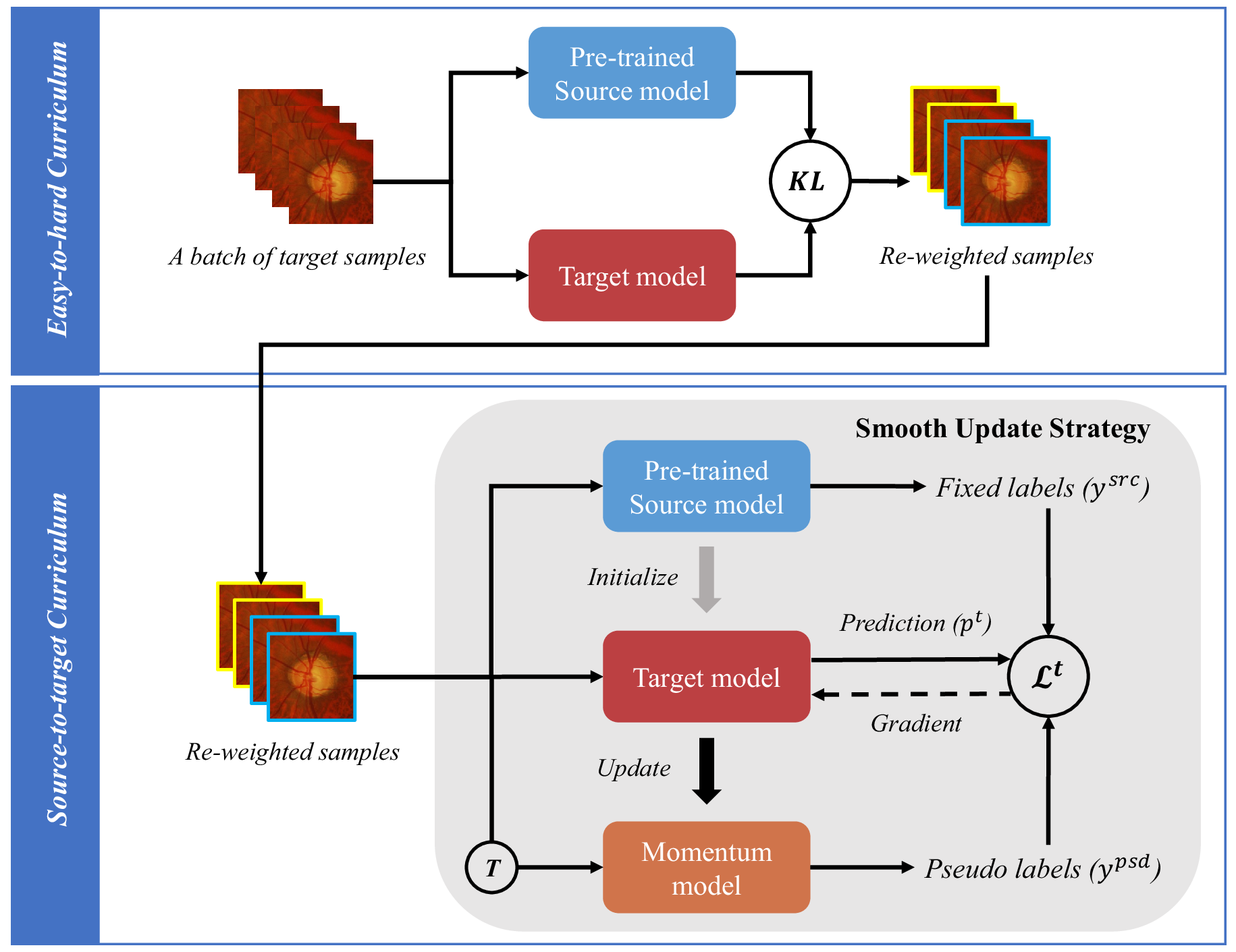}
    \caption{Pipeline of our learning from curriculum (LFC) framework. The proposed LFC consists of two curricula, \emph{i.e.,} easy-to-hard (top) and source-to-target (bottom).} \label{fig:framework}
\end{figure}

\section{Learning from Curriculum}
In this section, we first briefly introduce the paradigm of source-free domain adaptation, and then present the proposed learning from curriculum (LFC) framework in details. Under the setting of source-free domain adaptation, a model $f^s: \mathcal{X}^s \rightarrow \mathcal{Y}^s$ trained on unknown source domain $\mathcal{D}^s$ and a set of $N$ unlabeled target images $\mathcal{X}^t = \{x_1^t, \cdots, x_N^t\}$ are given. The aim of source-free UDA is to adapt the model $f^s$ using $\mathcal{X}^t$, and therefore obtain the target model $f^{t}$ with improved generalization on target domain.

\subsection{Overview} 
To achieve that, we propose a curriculum-based framework, as illustrated in Fig.~\ref{fig:framework}. First, the target images are sorted by an easy-to-hard curriculum based on the prediction consistency between the source and target models. 
The `easy' samples with the consistent predictions from the two models gain more attention from $f^t$ at the beginning of model adaptation, and the learning focus is gradually changed to the `hard' samples. The re-weighted target samples are then fed to a source-to-target curriculum, which is formed by a triplet-branch framework (\emph{i.e.,} source model $f^s$, target model $f^t$ and momentum model $f^m$). In this curriculum, we propose a smooth update strategy to enable the model start with $f^s$-generated pseudo labels and then gradually switch to the one generated by $f^t$. Such a progressive learning process can stablize the adaptation process and smoothly transfer the source-related knowledge to the target domain. In the following, we will introduce each curriculum in details.


\subsection{Easy-to-hard Curriculum}
Existing studies \cite{9461766,9392296,Wei_2021_WACV} have validated that a progressive learning procedure from `easy' to `hard' samples will, in fact, benefit feature representation learning during model adaptation. In the source-free UDA setting, however, only unlabeled target data is available. Thus, we cannot reveal the sample difficulty by explicitly measuring its distance to the two domains. In this regard, our easy-to-hard curriculum tries to measure the sample difficulty in the output space.

\vspace{2mm}
\noindent
{\bf Difficulty Measurement.} Given an input target image $x^t$, we first pass it through both $f^s$ and $f^t$ and obtain the predictive outputs $p^s$ and $p^t$, respectively (\emph{i.e.,} $p^s = f^s(x^t)$, $p^t = f^t(x^t)$). The $f^s$ is frozen and $f^t$ is initialized using AdaBN \cite{LI2018109} with $f^s$. Then, we calculate the Kullback-Leibler (KL) divergence $d$ between the two predictions:

\begin{equation}
    d = KL(p^s, p^t) = \frac{1}{W\times H}\sum_{u\in W, v\in H} p^s(u, v) log\frac{p^s(u, v)}{p^t(u, v)}, \label{eq:KL}
\end{equation}
where $W$ and $H$ are the width and height of the segmentation probability map, respectively; $(u, v)$ is the coordinate. The KL divergence represents the consistency between the predictions yielded by $f^s$ and $f^t$. Intuitively, the samples with consistent predictions fall in the overlapping area of source and target domains in the latent space, which result in a low value of $d$ (\emph{i.e.,} easy samples). The target model should start learning with these easy samples at the beginning of model adaptation and gradually change its focus to the `hard' ones.

\vspace{2mm}
\noindent{\bf Data Re-weighting.} In the proposed LFC, we re-weight the data in each training batch to dynamically tune the optimization direction. For a batch of target images $\mathcal{X}^t_B = \{x_b^t\}_{b=1}^B$, where $B$ is the batch size, the corresponding KL-divergence $d_b$ is first calculated for each $x_b^t$ using Eq.~\ref{eq:KL}, and then the sample weight $\omega$ is accordingly assigned:

\begin{equation}
    \omega_b = \alpha \left (\delta - \frac{d_b}{\sum_{b=1}^B d_b} \right ) + (1-\alpha),\label{eq2}
\end{equation}
where $\delta$ is a constant ($\delta$ is empirically set to $1.5$ in our experiments); $\alpha$ is a weight parameter adaptively tuning the learning focus:

\begin{equation}
    \begin{aligned}
        \alpha = 1 - sigmoid\left (\frac{R} {R_{max}}\right ),\label{eq:alpha}
    \end{aligned}
\end{equation}
where $R$ and $R_{max}$ are the current epoch and the number of pre-set tuning epochs ($R_{max}=5$ in the experiments), respectively.

\subsection{Source-to-target Curriculum}
Existing source-free UDA framework \cite{chen2021source} simply generated the pseudo labels $\mathcal{Y}^{src}\in \{y^{src}_1,\cdots, y^{src}_N\}$ with $f^s$. Since the $f^s$ is frozen, the pseudo labels cannot gain further updates during the process of model adaptation. To smoothly transfer source-related knowledge stored in $f^s$ to target domain, we propose a source-to-target curriculum, which enables the model to adaptively use the pseudo labels yielded by different models at different stages of model adaptation. As shown in Fig.~\ref{fig:framework}, in this curriculum, the re-weighted target samples $\omega D^t$ are fed to a triplet-branch framework, consisting of source model $f^s$, target model $f^t$ and momentum model $f^m$. Note that the three branches have the same backbone architecture. As mentioned before, the $f^s$ is frozen and $f^t$ is initialized using AdaBN \cite{LI2018109} with $f^s$. The parameters of $f^m$ are an exponential moving average of $f^t$:
\begin{equation}
  f^m \leftarrow \tau f^m+(1-\tau) f^t, \label{eq:EMA}
\end{equation}
where $\tau$ is the decay rate controlling the weight updating procedure. The triplet-branch framework is updated using our smooth update strategy incorporated with a self-supervised learning task.

\vspace{2mm}
\noindent{\bf Self-supervised Task: Learning with Augmentation Consistency.} To encourage the target model to deeply exploit transformation-invariant features from unlabeled target data, we randomly select an operation $T$ from a pre-defined transformation pool, consisting of cropping, vertical flipping and horizontal flipping, to transform the input target image $x^t$. The transformed image $T(x^t)$ is then sent to the momentum model $f^m$ for pseudo label $y^{psd}$ generation:

\begin{equation}
    y^{psd} = softmax(T^{-1}(f^m(T(x^t))),
\end{equation}
where $T^{-1}$ is the inverse transformation of $T$; $softmax(\cdot)$ is the softmax activation. After that, we enforce the consistency between $p^t=f^t(x^t)$ and $y^{psd}$ via cross-entropy loss:

\begin{equation}
    \mathcal{L}_{sl} = -\sum_{u\in W, v\in H} y^{psd}(u, v) log(p^t(u, v)).\label{eq:ce}
\end{equation}

\vspace{2mm}
\noindent{\bf Smooth Update Strategy.} At the beginning of model adaptation, the discrepancy between source and target domains may lead to sharp gradients, which may collapse the target model and degrade the quality of $\mathcal{Y}^{psd}\in \{y^{psd}_1,\cdots, y^{psd}_N\}$. Hence, our smooth update strategy starts the model learning with $\mathcal{Y}^{src}$, and then gradually varies the learning objective to $\mathcal{Y}^{psd}$ after several training epochs; therefore, the pseudo label $\mathcal{Y}^{psd}$ can gain continual updates across the whole process of modal adaptation. The cross-entropy loss (Eq.~\ref{eq:ce}) is also adopted for loss calculation with $\mathcal{Y}^{src}$ (denoted as $\mathcal{L}_{fix}$). Hence, the overall objective $\mathcal{L}^t$ for $f^t$ can be formulated as:

\begin{equation}
    \label{total loss}
    \mathcal{L}^t=\omega (\alpha \mathcal{L}_{fix}(p^t, y^{src})+ (1-\alpha)\mathcal{L}_{sl}(p^t, y^{psd})),
\end{equation}
where $\omega$ is the sample weight yielded by easy-to-hard curriculum (\emph{cf.} Eq.~\ref{eq2}); $\alpha$ is the weight parameter defined in Eq.~\ref{eq:alpha}.

\begin{table}[!b]
    \caption{Statistics of the retinal fundus image and colonoscopic image datasets.}\label{fundus-data}
    \centering
    \small
    \scalebox{1.}{
    \begin{tabular}{p{2.8cm}<{\centering}|p{1.8cm}<{\centering}|p{2.5cm}<{\centering}|p{3.3cm}<{\centering}}
    \hline
    {\bf Task} & {\bf Domain} & {\bf Dataset} & {\bf Number of Samples}\\
    \hline\hline
    \multicolumn{1}{c|}{\multirow{3}{*}{Fundus Segmentation}}& Source & REFUGE \cite{orlando2020refuge}  & 400 (Train)\\
    \cline{2-4}
    \multicolumn{1}{l|}{}   & Target & RIM-ONE-r3 \cite{fumero2011rim}   & 99 (Train) + 60 (Test)  \\
    \cline{2-4}
    \multicolumn{1}{l|}{}  & Target &  Drishti-GS \cite{sivaswamy2015comprehensive}   &  50 (Train) + 51 (Test) \\\hline
    \hline
    \multicolumn{1}{c|}{\multirow{2}{*}{Polyp Segmentation}}     & Source     & CVC-Clinic \cite{vazquez2017benchmark}  & 490 (Train)\\
    \cline{2-4}
    \multicolumn{1}{l|}{}   & Target & ETIS-Larib \cite{silva2014toward}        & 137 (Train) + 59 (Test)  \\
    \hline
    \end{tabular}}
\end{table}

\section{Experiments}
\begin{figure}[!h]
\centering
\subfigure{
\begin{minipage}[b]{\textwidth}
\includegraphics[width=1\textwidth]{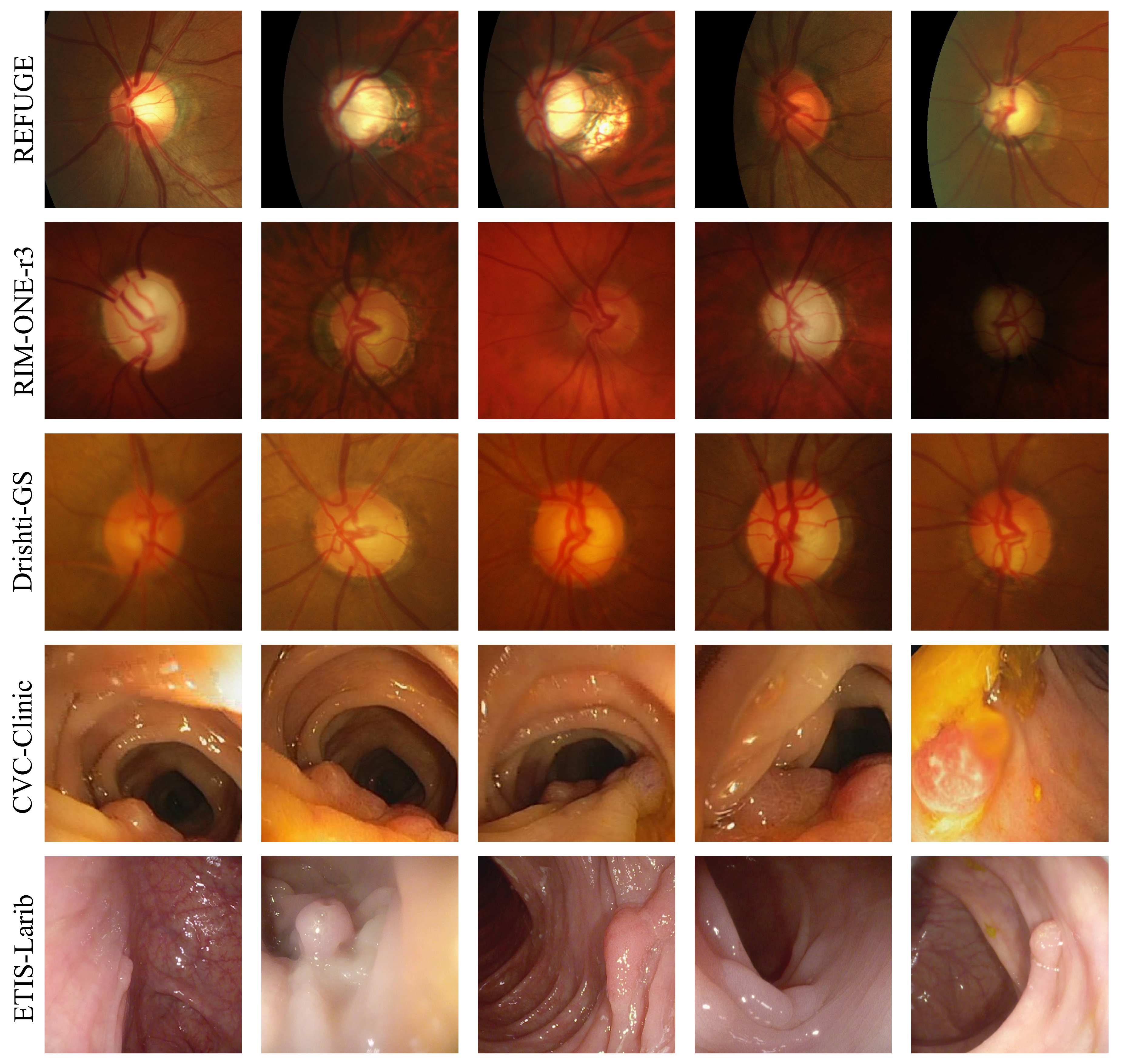}
\end{minipage}
}
\caption{Exemplars from fundus and colonoscopic image datasets adopted in our study. The source domains (\emph{i.e.,} REFUGE and CVC-Clinic) are visually different from the target ones (known as {\itshape domain shift}).}\label{exemplars}
\end{figure}

In this section, we evaluate the proposed curriculum-based framework on publicly available cross-domain datasets (detailed statistics of the datasets, including the number of samples and data separation, are listed in Table~\ref{fundus-data} and exemplars from each dataset are presented in Fig.~\ref{exemplars})
for fundus segmentation and polyp segmentation, respectively.  Also, an ablation study is conducted to validate the effectiveness of each curriculum in our framework.


\subsection{Datasets}

\vspace{2mm}
\noindent{\bf Fundus Segmentation.} We conduct experiments on retinal fundus images for optic disc (OD) and optic cup (OC) segmentation. Fundus image datasets are collected from different clinical centers. Specifically, the public training set with pixel-wise annotations of REFUGE \cite{orlando2020refuge} is adopted as the source domain, while the RIM-ONE-r3 \cite{fumero2011rim} and Drishti-GS \cite{sivaswamy2015comprehensive} datasets are treated as different target domains. The experimental settings, \emph{e.g.,} training/test separation, are consistent to \cite{chen2021source}. The vanilla UDA algorithms, consisting of the ones accessing source data (\emph{i.e.,} BEAL \cite{wang2019boundary} and AdvEnt \cite{vu2019advent}) and altering source domain training (\emph{i.e.,} SRDA \cite{bateson2020source} and DAE \cite{karani2021test}), are involved for comparison. The state-of-the-art source-free UDA approaches (\emph{i.e.,} DPL \cite{chen2021source} and SI \cite{SFD-MM}) are also evaluated.

\vspace{2mm}
\noindent{\bf Polyp Segmentation.} To further validate the effectiveness of our LFC, the experiments on colonoscopic images for polyp segmentation are also conducted. The publicly available colonoscopic image datasets, \emph{i.e.,} CVC-Clinic \cite{vazquez2017benchmark} and ETIS-Larib \cite{silva2014toward}, are used as source and target domains, respectively. In addition to DPL \cite{chen2021source}, the state-of-the-art source-accessible (IB-GAN \cite{Chen_TM_2021}) and source-free (FSM \cite{YANG2022102457}) UDA frameworks on this cross-domain task are also involved for comparison.

\subsection{Experimental Settings}

\vspace{2mm}
\noindent{\bf Implementation Details.} The proposed method is implemented using the PyTorch toolbox. We adopt two widely-used deep learning networks as backbone, \emph{i.e.,} DeepLab-V3 \cite{chen2017rethinking} and ResUNet-50 \cite{he2016deep,ronneberger2015u} for OC/OD segmentation and polyp segmentation, respectively, which are also employed by other comparison methods. The Adam optimizer (momentum=0.9) is adopted for network optimization with an initial learning rate of 0.001. The target model is observed to converge after ten epochs of tuning.

\vspace{2mm}
\noindent{\bf Evaluation Criteria.} Two commonly-used metrics are utilized for performance evaluation, \emph{i.e.,} Dice score and the average surface distance (ASD). Concretely, the former one measures the pixel-wise accuracy, while the latter one assesses the agreement of boundaries. The higher Dice means the better performance, which is opposite for ASD.

\subsection{Performance Evaluation}
In this section, we present the experimental results on the two tasks to validate the effectiveness of our LFC. 

\begin{table}[!t]
    \caption{Comparison of different methods on the target domain datasets for OC/OD segmentation. Note: {\em{S.}} denotes source domain; -- means the results are not reported by the methods. The segmentation accuracies of benchmarking frameworks in this table are duplicated from \cite{chen2021source}.}\label{result-fundus}
    \centering
    \small
    \resizebox{0.95\textwidth}{!}{
    \begin{tabular}{lccccc}
    \hline
    \multicolumn{1}{c|}{\multirow{2}{*}{Method}} & \multicolumn{1}{c|}{\multirow{2}{*}{S. Data Access}}                           & \multicolumn{2}{c|}{OD Segmentation}                                             & \multicolumn{2}{c}{OC Segmentation}                     \\ \cline{3-6} 
    \multicolumn{1}{c|}{}                 & \multicolumn{1}{c|}{} & \multicolumn{1}{c|}{Dice (\%)}            & \multicolumn{1}{c|}{ASD (pixel)}         & \multicolumn{1}{c|}{Dice (\%)}      & ASD (pixel)           \\ \hline\hline
    \multicolumn{6}{c}{\textbf{RIM-ONE-r3 \cite{fumero2011rim} }}                                                                                                                                                                                                                              \\ \hline
    \multicolumn{1}{l|}{w/o adaptation}        &       \multicolumn{1}{c|}{}                              & \multicolumn{1}{c|}{83.18±6.46}          & \multicolumn{1}{c|}{24.15±15.58}        & \multicolumn{1}{c|}{74.51±16.40}   & 14.44±11.27          \\
    \multicolumn{1}{l|}{Oracle \cite{wang2019boundary}}         &  \multicolumn{1}{c|}{}                                   & \multicolumn{1}{c|}{96.80}                & \multicolumn{1}{c|}{--}                  & \multicolumn{1}{c|}{85.60}          & --                    \\ \hline
    \multicolumn{1}{l|}{BEAL \cite{wang2019boundary}}       & \multicolumn{1}{c|}{\checkmark}                                       & \multicolumn{1}{c|}{89.80}                & \multicolumn{1}{c|}{--}                  & \multicolumn{1}{c|}{{81.00}}   & --                    \\
    \multicolumn{1}{l|}{AdvEnt \cite{vu2019advent}}       & \multicolumn{1}{c|}{\checkmark}                                   & \multicolumn{1}{c|}{89.73±3.66}          & \multicolumn{1}{c|}{9.84±3.86}          & \multicolumn{1}{c|}{77.99±21.08}   & \textbf{7.57±4.24}   \\
    \multicolumn{1}{l|}{SRDA \cite{bateson2020source}}                        & \multicolumn{1}{c|}{\checkmark}       & \multicolumn{1}{c|}{89.37±2.70}          & \multicolumn{1}{c|}{9.91±2.45}          & \multicolumn{1}{c|}{77.61±13.58}   & 10.15±5.75           \\
    \multicolumn{1}{l|}{DAE \cite{karani2021test}}                             & \multicolumn{1}{c|}{\checkmark}       & \multicolumn{1}{c|}{89.08±3.32}          & \multicolumn{1}{c|}{11.63±6.84}         & \multicolumn{1}{c|}{79.01±12.82}   & 10.31±8.45           \\ \hline
    \multicolumn{1}{l|}{DPL \cite{chen2021source}} &      \multicolumn{1}{c|}{}                      & \multicolumn{1}{c|}{90.13±3.06}          & \multicolumn{1}{c|}{{9.43±3.46}}          & \multicolumn{1}{c|}{79.78±11.05}   & 9.01±5.59           \\
    \multicolumn{1}{l|}{DPL$^*$ \cite{chen2021source}} &      \multicolumn{1}{c|}{}                 & \multicolumn{1}{c|}{89.96±3.74}          & \multicolumn{1}{c|}{{9.54±3.83}} & \multicolumn{1}{c|}{78.82±12.23}   & 9.14±6.11            \\
    \multicolumn{1}{l|}{SI \cite{SFD-MM}} &    \multicolumn{1}{c|}{}                      & \multicolumn{1}{c|}{91.43±1.74}          & \multicolumn{1}{c|}{\textbf{9.22±3.44}}          & \multicolumn{1}{c|}{80.99±1.80}   & 10.54±4.46           \\
    \multicolumn{1}{l|}{LFC (\itshape ours)}               &             \multicolumn{1}{c|}{}               & \multicolumn{1}{c|}{\textbf{91.87±1.64}} & \multicolumn{1}{c|}{9.39±3.73}          & \multicolumn{1}{c|}{\textbf{81.27±11.25}}   & 9.06±5.86            \\ \hline
    \hline
    \multicolumn{6}{c}{\textbf{Drishti-GS \cite{sivaswamy2015comprehensive} }}                                                                                                                                                                                                                              \\ \hline
    \multicolumn{1}{l|}{w/o adaptation}      &      \multicolumn{1}{c|}{}                              & \multicolumn{1}{c|}{93.84±2.91}          & \multicolumn{1}{c|}{9.05±7.50}          & \multicolumn{1}{c|}{83.36±11.95}   & 11.39±6.30           \\
    \multicolumn{1}{l|}{Oracle \cite{wang2019boundary}}      &   \multicolumn{1}{c|}{}                            & \multicolumn{1}{c|}{97.40}                & \multicolumn{1}{c|}{--}                  & \multicolumn{1}{c|}{90.10}          & --                    \\ \hline
    \multicolumn{1}{l|}{BEAL \cite{wang2019boundary}}       & \multicolumn{1}{c|}{\checkmark}                                    & \multicolumn{1}{c|}{96.10}                & \multicolumn{1}{c|}{--}                  & \multicolumn{1}{c|}{{86.20}} & --                    \\
    \multicolumn{1}{l|}{AdvEnt \cite{vu2019advent}}         & \multicolumn{1}{c|}{\checkmark}                               & \multicolumn{1}{c|}{96.16±1.65}          & \multicolumn{1}{c|}{4.36±1.83}          & \multicolumn{1}{c|}{82.75±11.08}   & 11.36±7.22           \\
    \multicolumn{1}{l|}{SRDA \cite{bateson2020source}}                               & \multicolumn{1}{c|}{\checkmark}       & \multicolumn{1}{c|}{96.22±1.30}          & \multicolumn{1}{c|}{4.88±3.47}          & \multicolumn{1}{c|}{80.67±11.78}   & 13.12±6.48           \\
    \multicolumn{1}{l|}{DAE \cite{karani2021test}}                                      & \multicolumn{1}{c|}{\checkmark}       & \multicolumn{1}{c|}{94.04±2.85}          & \multicolumn{1}{c|}{8.79±7.45}          & \multicolumn{1}{c|}{83.11±11.89}   & 11.56±6.32           \\ \hline
    \multicolumn{1}{l|}{DPL \cite{chen2021source}} &      \multicolumn{1}{c|}{}                      & \multicolumn{1}{c|}{96.39±1.33}          & \multicolumn{1}{c|}{\textbf{4.08±1.49}}          & \multicolumn{1}{c|}{83.11±17.80}   & 11.39±10.18           \\
    \multicolumn{1}{l|}{DPL$^*$ \cite{chen2021source}} &    \multicolumn{1}{c|}{}                      & \multicolumn{1}{c|}{96.05±1.28}          & \multicolumn{1}{c|}{4.18±1.55}          & \multicolumn{1}{c|}{82.84±17.21}   & 11.38±9.89           \\
    \multicolumn{1}{l|}{SI \cite{SFD-MM}} &    \multicolumn{1}{c|}{}                      & \multicolumn{1}{c|}{96.86±1.03}          & \multicolumn{1}{c|}{6.02±3.66}          & \multicolumn{1}{c|}{85.96±0.97}   & \textbf{11.20±6.56}           \\
    \multicolumn{1}{l|}{LFC ({\itshape ours})}               &   \multicolumn{1}{c|}{}                            & \multicolumn{1}{c|}{\textbf{97.04±1.01}} & \multicolumn{1}{c|}{{4.09±1.48}} & \multicolumn{1}{c|}{\textbf{86.31±12.11}}   & {11.27±10.16} \\ \hline
    \multicolumn{6}{l}{\scriptsize * For a fair comparison, we re-implement DPL using the same $f^s$ adopted by our LFC.}
    \end{tabular}}
\end{table}

\vspace{2mm}
\noindent{\bf Fundus Segmentation.} Consistent to \cite{chen2021source}, we present the results of `w/o adaptation' as the lower bound and `Oracle' (fully-supervised training with target data) as the upper bound.
The evaluation results of benchmarking methods and our LFC are shown in Table \ref{result-fundus}. 
It can be easily observed that all UDA methods can improve the segmentation accuracy on target domain over the baseline (w/o adaptation) to some extent. Specifically, our LFC achieves the Dice scores of $91.87\%$ and $97.04\%$ for OD segmentation on the target domains, respectively, which even outperform the best source-accessible UDA approach (BEAL). Such experimental results demonstrate the potential of source-free UDA setting for practical applications. Furthermore, the proposed LFC achieves the consistent higher accuracies for OD and OC segmentation compared with the state-of-the-art source-free UDA framework (DPL) on both target domains.
The underlying reason may be that DPL treats all samples equally important and uses the pseudo labels generated by source model across the whole adaptation process. Such a fixed learning process narrows the space for model to adaptively tune its learning focus. In contrast, the two proposed curricula make the whole adaptation process progressive and flexible, which boost the segmentation accuracy.

\vspace{2mm}
\noindent{\bf Polyp Segmentation.} We also evaluate our method on colonoscopic datasets for polyp segmentation. The evaluation results of state-of-the-art methods (\emph{i.e.,} DPL \cite{chen2021source} and FSM \cite{YANG2022102457}) and our LFC are shown in Table \ref{result-colon}. The source-accessible UDA framework (\emph{i.e.,} IB-GAN) achieves the best Dice source of $73.01\%$ on the target domain. In comparison, our LFC achieves the comparable Dice score ($69.23\%$) without access of the source data, which is $+1.13\%$ higher than the runner-up (FSM). It is worthwhile to mention that the performance of FSM is different from the one reported in \cite{YANG2022102457}. The reason is that we used CVC-Clinic, a subset of EndoScene, to train the source model for FSM in the experiment, while Yang \emph{et al.} \cite{YANG2022102457} adopted the whole EndoScene dataset, consisting of 912 samples, for source model training.

\begin{table}[!t]
    \caption{Performance of polyp segmentation on the ETIS-Larib dataset \cite{vazquez2017benchmark}.}\label{result-colon}
    \centering
    \footnotesize
    \resizebox{0.95\textwidth}{!}{
    \begin{tabular}{l|p{2.5cm}<{\centering}|p{2cm}<{\centering}|p{2cm}<{\centering}p{2cm}<{\centering}p{2cm}<{\centering}}
    \hline
                    &  \multirow{2}{*}{w/o adaptation}          &    \multirow{2}{*}{IB-GAN \cite{Chen_TM_2021}}   & \multicolumn{3}{c}{\bf Source-free}          
                    \\\cline{4-6}   
                    &                              &    &  DPL \cite{chen2021source}    &   FSM \cite{YANG2022102457} & LFC ({\itshape ours})                                       \\\hline\hline
    Dice (\%)       &  {64.46±3.44}             &   73.01±1.52   &  {67.14±2.81}        & {68.69±2.67}              & \textbf{69.82±2.73}            \\\hline
    ASD  (pixel)    &  {18.02±7.13}             &   10.18±4.74   &  {11.92±4.13}        & {11.81±3.51}              & \textbf{11.78±3.82}          \\\hline
    \end{tabular}}
\end{table}

\begin{table}[!t]
\caption{Ablation study of LFC on the RIM-ONE-r3 dataset \cite{fumero2011rim}.}\label{ablation}
\centering
\small
\resizebox{0.95\textwidth}{!}{
\begin{tabular}{l|cc|cc}
\hline
\multicolumn{1}{c|}{\multirow{2}{*}{}}                & \multicolumn{2}{c|}{OD   Segmentation}                 & \multicolumn{2}{c}{OC   Segmentation}          \\ \cline{2-5} 
                                        & \multicolumn{1}{c|}{Dice (\%)}                          & ASD (pixel)  & \multicolumn{1}{c|}{Dice (\%)}    & ASD (pixel)  \\ \hline\hline
w/o adaptation                          & \multicolumn{1}{c|}{83.18±6.46}                        & 24.15±15.58 & \multicolumn{1}{c|}{74.51±16.40} & 14.44±11.27 \\ 
{\itshape Ours} w/o easy-to-hard        & \multicolumn{1}{c|}{89.05±2.84}                        & 9.81±5.12   & \multicolumn{1}{c|}{79.33±10.21} & 10.13±6.07  \\ 
{\itshape Ours} w/o source-to-target    & \multicolumn{1}{c|}{88.42±3.75}                        & {9.66±4.26}   & \multicolumn{1}{c|}{78.62±11.76} & 9.29±6.41   \\ 
LFC (\itshape ours)                     & \multicolumn{1}{c|}{\textbf{91.87±1.64}} & \textbf{9.39±3.73}   & \multicolumn{1}{c|}{\textbf{81.27±11.25}} & \textbf{9.06±5.86}   \\ \hline
\end{tabular}}
\end{table}

\subsection{Ablation Study} 
The proposed curriculum-based framework consists of two curricula---easy-to-hard and source-to-target. To validate the contribution made by each curriculum, we conduct an ablation study on the RIM-ONE-r3 dataset and present the results in Table~\ref{ablation}. As shown, removing either curriculum, the Dice scores for OC/OD segmentation significantly degrade. On the other hand, the best Dice scores, \emph{i.e.,} $91.87\%$ and $81.27\%$ for OD and OC, respectively, are achieved by jointly using the easy-to-hard and source-to-target curricula. 

\section{Conclusion}
In this paper, we proposed a curriculum-based framework, namely learning from curriculum (LFC), for source-free domain adaptation, which consists of two curricula---easy-to-hard and source-to-target. 
The former curriculum enables the framework to start learning with `easy' samples and gradually tune the optimization direction of model adaption by increasing the sample difficulty. While, the latter one can stablize the adaptation process, which ensures model's smooth transfer from source domain to the target.
Extensive experiments on publicly available cross-domain datasets for fundus segmentation and polyp segmentation showed that our framework surpassed existing approaches and achieved a new state-of-the-art.

\bibliographystyle{elsarticle-num} 
\bibliography{reference}

\end{document}